\documentclass{foundationpaper}

\usepackage{booktabs}
\usepackage{enumitem}
\usepackage{pifont}
\usepackage{float}
\newcommand{\full}{\ding{51}}
\newcommand{\pmark}{\raisebox{0.2ex}{\small$\circ$}}
\newcommand{\none}{--}

\hypersetup{
    pdftitle={Foundation Protocol},
    pdfauthor={FoundationAgents Team},
    pdfsubject={A Coordination Layer for Agentic Society}
}

\title{\centering Foundation Protocol: \\ A Coordination Layer for Agentic Society}

\author[1,2,*,\dagger]{Bang Liu}
\author[1,*]{Yongfeng Gu}
\author[1,4,*]{Jiayi Zhang}
\author[1,*]{Zhaoyang Yu}
\author[1,*]{Sirui Hong}
\author[5,*]{Maojia Song}
\author[1,2,*]{Xiaoqiang Wang}
\author[1]{Mingyi Deng}
\author[1]{Zijie Zhuang}
\author[1]{Ronghao Wang}
\author[1]{Mingzhe Cao}
\author[1]{Yutong Zhu}
\author[1]{Xingjian Li}
\author[1]{Yifan Wu}
\author[1]{Jianhao Ruan}
\author[1]{Yiran Peng}
\author[1]{Shuangrui Chen}
\author[1]{Jinlin Wang}
\author[1]{Yizhang Lin}
\author[1]{Dongjie Zhang}
\author[1,2]{Dekun Wu}
\author[6]{Chen Ma}
\author[7]{Lizi Liao}
\author[8]{Han Yu}
\author[9]{Jian Pei}
\author[10]{Heng Ji}
\author[11]{Qiang Yang}
\author[1,4,\dagger]{Yuyu Luo}
\author[1,3,\dagger]{Chenglin Wu}

\affiliation[1]{FoundationAgents}
\affiliation[2]{Universit\'{e} de Montr\'{e}al \& Mila}
\affiliation[3]{DeepWisdom}
\affiliation[4]{HKUST(GZ)}
\affiliation[5]{Singapore University of Technology and Design}
\affiliation[6]{City University of Hong Kong}
\affiliation[7]{Singapore Management University}
\affiliation[8]{Nanyang Technological University}
\affiliation[9]{Duke University}
\affiliation[10]{University of Illinois Urbana-Champaign}
\affiliation[11]{Hong Kong Polytechnic University}

\contribution[*]{Core contributors.}
\contribution[\dagger]{Corresponding authors.}

\abstract{Autonomous agents are moving from tools into a layer of social infrastructure: they browse, purchase, deploy software, manage systems, and increasingly interact with one another. As these systems scale, the bottleneck shifts away from raw model capability toward coordination. Agents need to form reliable relationships, organize multi-agent work, exchange value, support an AI economy, and stay safe and accountable under real-world oversight.
This paper introduces the \emph{Foundation Protocol (FP)}, a graph-first coordination layer for an emerging human--AI society. FP unifies heterogeneous entities, including agents, tools, resources, humans, institutions, and organizations, and supports native multi-party organization and event-based collaboration. It also provides economic primitives for metering, receipts, and settlement, and treats policy, provenance, and audit as first-class concerns. FP is designed to wrap and bridge existing protocols rather than replace them, enabling incremental adoption while reducing integration and governance overhead.
The aim is to keep autonomous agency composable while keeping accountability non-negotiable, so that coordination itself can become shared infrastructure for a human–AI society that is open, pluralistic, and governable.}

\metadata[Keywords]{Foundation Protocol, Agentic Society, Coordination Protocols, Multi-Agent Systems, AI Economy, Trust and Identity, Governance and Auditability}
\metadata[Project]{\url{https://github.com/FoundationAgents/foundation-protocol}}

\begin{document}

\maketitle

\section{Introduction}
\label{sec:intro}

Autonomous agents are beginning to enter the internet not just as tools we operate, but as participants that can act on our behalf. They read and write to the same services we do, hold long-lived credentials, purchase resources, and deploy software.
Their decisions carry financial, operational, and reputational consequences.
In early deployments, an agent may be little more than a thin natural-language layer over a few APIs. More ambitious systems use it as a persistent operator: one that plans, coordinates, negotiates, and acts across services over time.
This shift changes the role of protocols. Protocols are the agreements that make such systems interoperable. They are not libraries or SDKs, but shared choreographies: which roles exist, what messages mean, what authority is delegated, and which state transitions are allowed. For agentic systems, this boundary matters because communication often \emph{is} execution, and execution carries economic, social, and governance consequences.

Consider what an ordinary workflow looks like once agents become common. A user might ask a personal agent to arrange travel across several vendors, negotiate refunds, and stay within a budget. To do this, the agent recruits specialist agents for itinerary planning, price monitoring, policy compliance, and payment execution. At sensitive points, it asks the user for approval. When the trip is over, it settles with service providers through auditable receipts.
The same pattern soon extends beyond personal assistance. 
A research group may assemble an AI team to search the literature, rent GPU time, coordinate instruments, run analyses, and produce a provenance trail that withstands later review. 
A one-person company may operate through a network of agents that handle design, engineering, procurement, compliance, sales, and customer support. 
In more autonomous settings, AI organizations may form and dissolve. They hire external services, compete for resources, and interact with human institutions under explicit rules.
The important point is not that these examples belong to different domains. It is that they share the same structure. Agents, humans, tools, services, companies, and institutions become nodes in an evolving graph. They delegate authority, form teams, exchange value, enforce policy, and leave evidence behind. What we are describing is not one conversation, nor even a single multi-agent chat. It is a hybrid human--AI society in miniature. Once agents recruit, transact, report, and act across organizational boundaries, identity, budget, provenance, and oversight can no longer be added as afterthoughts. They become part of the communication substrate itself.

Early systems already show fragments of this pattern. OpenClaw presents a locally run, chat-controlled agent runtime. It can sit inside ordinary communication channels and coordinate tool use through an expanding skills ecosystem \cite{openclaw_site,openclaw_github}. Moltbook pushes the idea in a different direction. It is a social layer where agents maintain profiles, post updates, authenticate one another, and interact while humans observe from the outside \cite{moltbook_site,wired_moltbook}. They show the same shift from different sides. Agents are no longer only interfaces to tools. They are becoming persistent entities that communicate, delegate, and encounter other entities in shared environments.
This trend also changes what ``communication'' means. In conventional software, a message usually carries information. In agentic systems, a message may trigger code execution, resource use, payment, delegation, or policy change. The boundary between communication and execution becomes thin when an agent ingests untrusted content, downloads third-party code, and acts with persistent credentials. Microsoft's security research team describes self-hosted agent runtimes as untrusted code execution with durable privileges, and recommends isolation, scoped identity, and continuous monitoring \cite{microsoft_openclaw_security}. Autonomy therefore turns the protocol layer into a safety boundary. The protocol is no longer only an integration convenience. It is where the system records identity, delegates authority, leaves evidence, and enforces accountability.

The existing protocol already covers several important pieces of agent interaction. MCP gives models a common way to use tools \cite{mcp_spec}. A2A defines a surface for agent-to-agent task collaboration \cite{a2a_spec}. A2UI focuses on controllable delegation through user interfaces \cite{a2ui_spec}. DIDComm provides secure DID-based messaging \cite{didcomm_spec}. ANP emphasizes discovery and negotiation among agents in open networks \cite{anp_paper}. and UCP targets commerce among autonomous participants \cite{ucp_spec}. Each addresses a real boundary. The problem is that an agentic society does not stay inside those boundaries. A single workflow may need tool use, agent delegation, UI control, identity verification, payment, policy enforcement, and audit across the same chain of action.

Fragmentation becomes costly here. When every protocol carries its own notion of identity, session state, authority, trace, and evidence, integration takes more than adapters. Semantics start to drift across layers. Provenance can break at protocol boundaries. Oversight becomes a patchwork of logs, receipts, access-control rules, and prompt fragments. Recent surveys of agent protocols point to related gaps around collaboration, scalability, security, privacy, and group-based interaction \cite{yang2025survey}. For FP, these gaps are not peripheral. They become central once autonomous entities form teams, exchange value, and operate under real-world accountability.

The consequence is both technical and institutional. If interoperability remains painful, vertical integration becomes the easiest path. A few platforms own identity, policy, routing, memory, and economic settlement end to end. If interoperability is improvised, open networks may still emerge, but they remain fragile, hard to audit, and difficult to defend against abuse. A foundation layer should avoid this false choice. It should make heterogeneous protocols easier to compose, and keep the important questions visible across the system, such as identity, authority, value, provenance, and governance.

This is the role of the \emph{Foundation Protocol (FP)}. FP is a graph-native protocol for heterogeneous agentic organizations, where coordination, economic exchange, and accountable execution share the same foundation layer. It treats agents, tools, resources, humans, institutions, and organizations as addressable entities in a shared graph. It represents relationships, memberships, sessions, and activities as first-class protocol objects. And it gives value exchange, policy, provenance, and audit a common evidence spine.
Its purpose is not to replace existing protocols. It is to provide the control-plane substrate that lets them compose across boundaries while preserving the identity, authority, and accountability needed for systems to remain governable as they scale.

\subsection{From Steam to Agents: Industrial Revolutions as Rises in Intelligence Density}
\label{subsec:intel-density}

An instructive way to read two centuries of industrial change is not only through machines or fuels, but through the density with which a society can gather and coordinate intelligence. By \emph{intelligence density}, we mean the amount of useful cognitive work that can be brought together within a social or technical system: how much know-how is available, how quickly it circulates, and how effectively it can be organized into action. Viewed through this lens, each industrial wave coincided with a step change in our capacity to aggregate and direct human knowledge. Steam and mechanization moved craft into organized production. Electricity and the assembly line professionalized engineering and industrial R\&D. Electronics and computing expanded the knowledge workforce. Industry~4.0 then fused networks, sensors, and cyber--physical feedback loops \cite{schwab2016fourth,hermann2016industrie}. These shifts were not merely technical. They also reorganized institutions, standards, finance, and production into new techno-economic paradigms, allowing knowledge to circulate and compound more efficiently \cite{perez2002technological}. Compatibility and network effects then accelerated this process further\cite{katz1985network}.

\begin{table}[ht]
\centering
\caption{Industrial revolutions through the lens of \emph{intelligence density}}
\label{tab:intel-density}
\footnotesize
\begin{tabular}{@{}p{0.23\textwidth} p{0.3\textwidth} p{0.44\textwidth}@{}}
\toprule
\textbf{Industrial Revolution} & \textbf{Technology drivers} & \textbf{Signature of intelligence density}\\
\midrule
First (c.\,1760--1840) & Steam engine, mechanized textiles, rail & Know-how concentrated in a small cadre of engineers and inventors; craft labor begins to organize around factories. \\
Second (c.\,1870--1914) & Electricity, internal combustion, chemistry, assembly line & Formal R\&D appears; engineering education scales; industrial laboratories and technical standards emerge, and knowledge becomes systematic.\\
Third (c.\,1940s--2000s) & Electronics, computing, automation, the internet & Knowledge workers proliferate; programmers, engineers, and scientists become core to production; global information flows. \\
Fourth (c.\,2010--) & AI, data, IoT, cyber--physical systems, cloud/edge & Cross-disciplinary fusion; platformized collaboration; tight sensor--software feedback loops in industry. \\
\bottomrule
\end{tabular}
\end{table}

Seen this way, the next step is already visible. While the fourth industrial revolution digitized processes, the next phase will systematize coordination among intelligent actors, both human and artificial. Agents provide reusable cognitive units; what remains missing is a common substrate through which these actors can discover one another, establish identity, form teams, exchange bounded context, transact, and leave auditable evidence across organizational boundaries. A foundation protocol determines whether this coordination becomes low-cost, open, and governable, or brittle, proprietary, and concentrated.

\subsection{From Hyperlinks to Hyperrealities: The Evolution and Lessons of Our Digital Society}
\label{subsec:web-evolve}

To see what a new foundation layer should preserve, repair, and extend, it is useful to look back at how the web evolved. Web~1.0 linked documents into a global information commons \cite{bernerslee1991proposal}. Web~2.0 turned readers into participants, but also concentrated power in surveillance-driven platforms \cite{oreilly2005web2,zuboff2019age}. Web~3.0 sought decentralization through cryptography and smart contracts, yet often struggled with fragmentation and usability \cite{buterin2014nextgen}. The next phase, sometimes described as an agentic or symbiotic web, adds pervasive AI, ambient computation, and mixed reality. Digital systems no longer only present information; they increasingly act, decide, and mediate relationships on our behalf.

\begin{figure}[!ht]
    \centering
    \includegraphics[width=\textwidth]{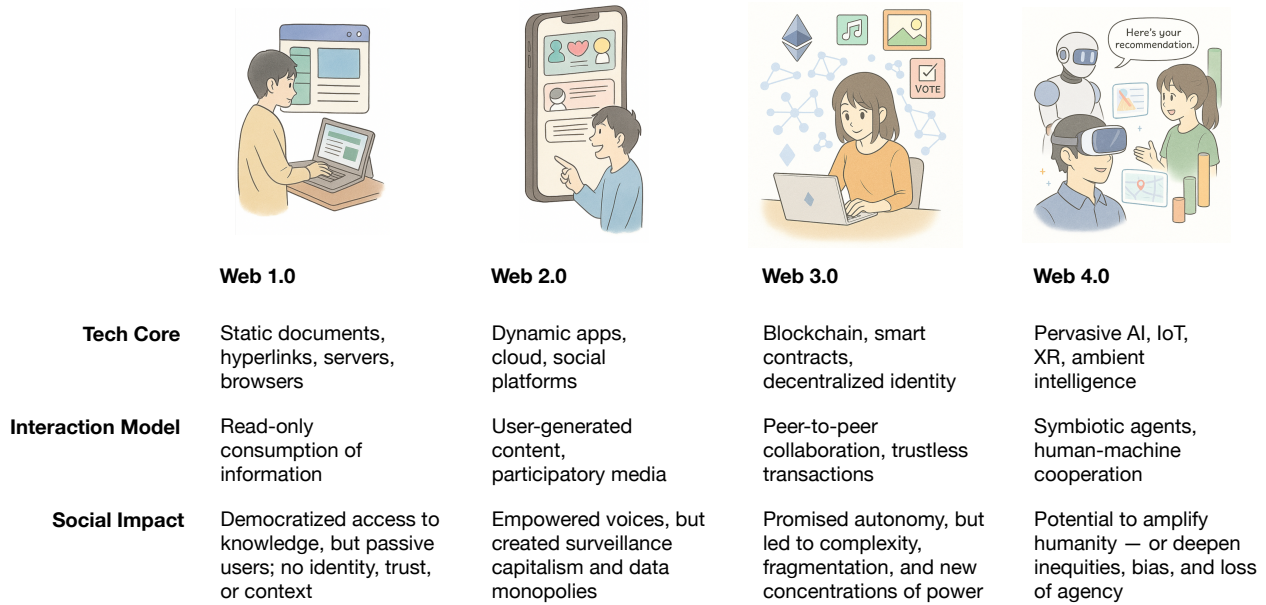}
    \caption{A compact view of the web's evolution: each generation raised capability while exposing new coordination failures. Web~4.0-like systems intensify those failures because agents act, interact, and transact at scale.}
    \label{fig:web-evolution}
\end{figure}

Figure~\ref{fig:web-evolution} compresses this history into a single view. Each generation expanded what the web could do, but also revealed a new coordination problem. Agentic systems make this problem sharper because they do not merely publish or consume content. They act, interact, and transact at scale.

Two lessons matter for FP. First, capability tends to arrive before the coordination primitives needed to govern it. The internet is very good at moving packets and linking resources; it is much less good at making clear who is acting, what authority has been delegated, what a message commits to, and who can be held accountable afterward. Second, as systems become more agentic, safety cannot remain outside the protocol layer. The web's next phase will not only distribute content. It will distribute agency. Once agency is distributed, identity, policy, provenance, and governance become part of the communication substrate itself.

\subsection{Design Objectives and the Case for a Foundation Layer}
\label{subsec:objectives}

A protocol for an agentic society is not defined by a single message type. Instead, it is determined by the ease, safety and cost at which all participants operate. A useful starting point is with \emph{behavioral closure}: what do autonomous agents need to do together when they share a world? In practice, most agentic systems repeatedly converge on four basic intents. They exchange information, coordinate work, exchange value for resources and services, and negotiate when preferences, constraints, or obligations conflict.

Existing protocols cover important parts of this space. MCP provides a strong interface for model-to-tool access; A2A offers a practical surface for agent-to-agent task collaboration; A2UI focuses on controllable interface delegation; DIDComm provides secure DID-based messaging; ANP emphasizes discovery and negotiation in open agent networks; and UCP targets agentic commerce \cite{mcp_spec,a2a_spec,a2ui_spec,didcomm_spec,anp_paper,ucp_spec}. Each addresses a real boundary. What remains under-specified is the shared substrate that these ecosystems repeatedly re-create in different forms: a unified notion of \emph{entity}, first-class \emph{organizations} beyond point-to-point sessions, interoperable \emph{economic attestations}, and an end-to-end \emph{evidence spine} suitable for audit and oversight.

FP's design objectives follow from this gap, and from the emerging economics of verification in autonomous systems \cite{virtual_agent_economies}. 
Recent economic analyses make this pressure clear. As autonomous execution becomes cheaper, the scarce complement shifts toward verification capacity, cryptographic provenance, and liability underwriting \cite{agi_economics}.
FP unifies heterogeneous entities under one addressable model and treats organizations, roles, and delegation as protocol primitives rather than middleware conventions. It structures interaction as events and streams with ordering and correlation, so collaboration remains observable as it scales. It adds economy primitives, including metering, receipts, settlement references, and dispute signals, in a ledger-agnostic form, so value exchange can be audited without mandating a payment rail. Finally, it makes governance first-class through policy enforcement points and provenance hooks, enabling systems where fast execution does not imply fragile accountability.

Two additional constraints shape the design. First, FP is built for \emph{progressive disclosure}: counterparts exchange minimal metadata by default and reveal detail on demand, reducing token and context overhead compared with the common pattern of copying full tool descriptions into a working prompt. Second, FP keeps its core small and moves variability into profiles, extensions, and bridges, enabling incremental adoption rather than a flag-day migration.

\begin{table}[!ht]
\centering
\caption{Where FP sits relative to widely-used protocols (\full: core focus, \pmark: partial/indirect, \none: out of scope).}
\label{tab:comparison}
\footnotesize
\begin{tabular}{@{}p{0.34\textwidth}ccccccc@{}}
\toprule
\textbf{Capability} & \textbf{FP} & \textbf{MCP} & \textbf{A2A} & \textbf{A2UI} & \textbf{DIDComm} & \textbf{ANP} & \textbf{UCP}\\
\midrule
Unified entities (agent/tool/human/org) & \full & \pmark & \pmark & \none & \pmark & \pmark & \pmark \\
Native groups and org structure & \full & \none & \pmark & \none & \none & \pmark & \none \\
Economy primitives (receipt/settle/dispute) & \full & \none & \none & \none & \none & \none & \pmark \\
Policy, audit, provenance as first-class & \full & \pmark & \pmark & \none & \pmark & \pmark & \pmark \\
Progressive disclosure (low overhead) & \full & \pmark & \pmark & \none & \none & \pmark & \pmark \\
Primary scope & society & tools & tasks & UI & secure msg & agentic web & commerce \\
\bottomrule
\end{tabular}
\end{table}

Table~\ref{tab:comparison} is not a scoreboard. It is a boundary map. FP is meant to complement these efforts by standardizing the cross-cutting substrate they inevitably share, while leaving domain-specific semantics to the protocols that already do them well. To keep the white paper focused on protocol essentials rather than comparisons between stacks, the appendix describes the reference implementation's architecture, core concepts, and key technical choices.

\subsection{Scope, Non-goals, and Paper Roadmap}
\label{subsec:scope}

FP is a coordination layer for agentic society, not an agent runtime or orchestration system. Its core standardizes how entities describe themselves, how multi-party interactions are formed and traced, how value exchange is metered and attested, and how policies and evidence remain coherent across organizational and protocol boundaries. FP does not prescribe a scheduler, a transport stack, an identity method, or a payment rail. Those choices belong to profiles, implementations, and deployment environments.

The rest of this paper is organized as follows. Section~\ref{sec:arch} introduces FP's plane-based architecture. Section~\ref{sec:scenarios} illustrates the kinds of systems this architecture is meant to support, first through a high-level survey of application categories and then through a detailed scenario that exercises every plane. Appendix~\ref{app:ref} describes the reference implementation's architecture, core concepts, and technical choices.

\section{The Architecture of the Foundation Protocol}
\label{sec:arch}

FP adopts a graph-native view of agentic systems. Entities are nodes; relationships, memberships, and sessions are edges; interactions are activities over the graph; and policy, provenance, and audit provide the evidence needed to govern those activities. This view leads to a plane-based architecture that keeps the protocol core small while making its extension points explicit.

\begin{figure}[!ht]
    \centering
    \includegraphics[width=\textwidth]{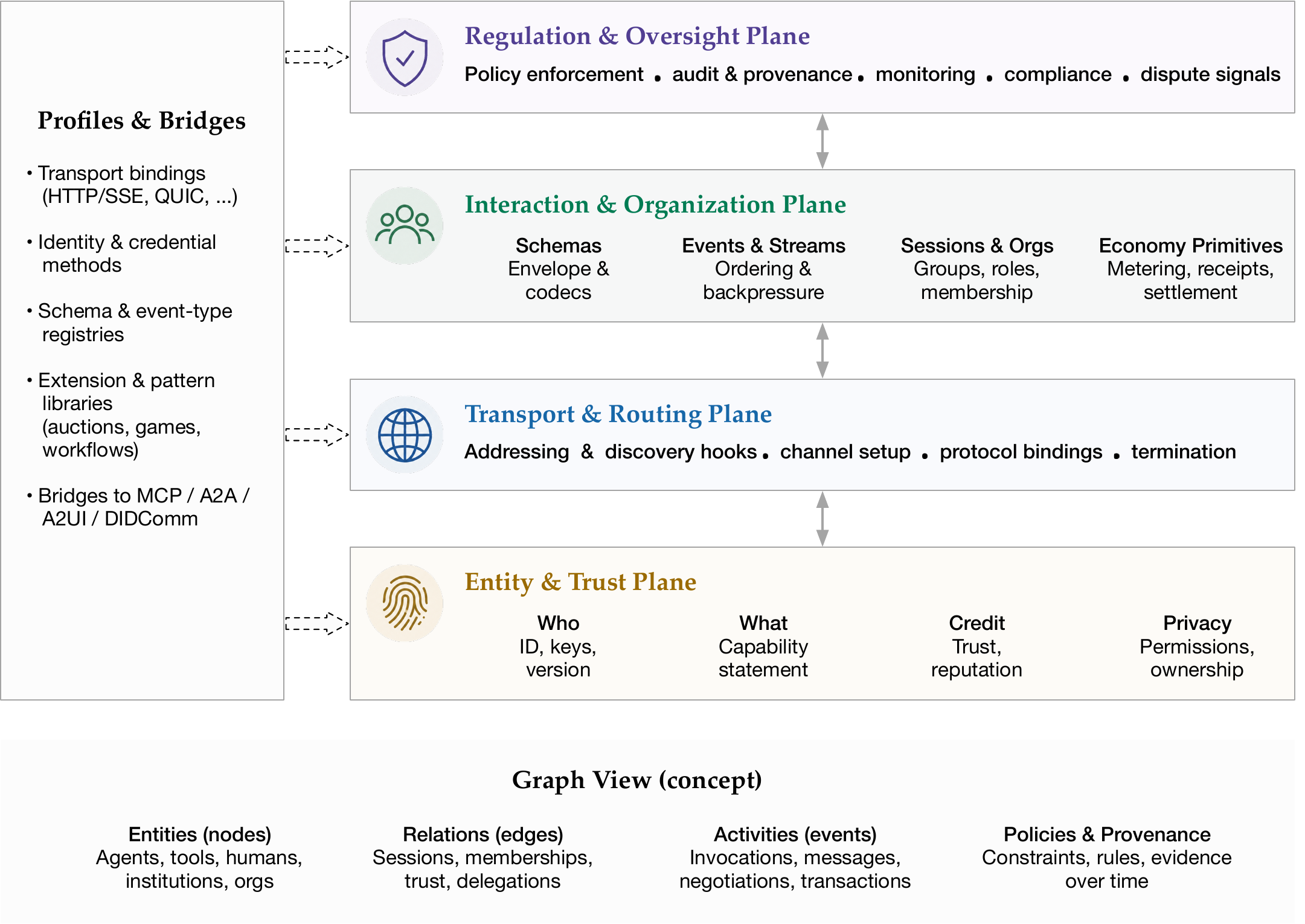}
    \caption{Foundation Protocol architecture (core): four planes plus a configuration/profile plane. The design keeps the core semantic commitments stable while allowing transport, identity, and domain-specific patterns to evolve independently.}
    \label{fig:fp-arch}
\end{figure}

Figure~\ref{fig:fp-arch} summarizes the FP core as four planes, with a separate configuration and profile plane that binds the core to concrete transports, identity methods, and extensions.
Each plane corresponds to a different kind of structure in the graph. The Entity \& Trust Plane defines the facts that make a node recognizable and accountable: identity, capabilities, credentials, trust signals, and privacy constraints. The Transport \& Routing Plane specifies how entities are addressed, discovered, connected, and reached across concrete transports. The Interaction \& Organization Plane defines the activities that occur among entities, from messaging and event streams to groups, roles, transactions, and settlements. The Regulation \& Oversight Plane provides the policy and evidence layer through which these activities can be monitored, constrained, reviewed, and audited as systems scale.

\subsection{A Minimal Vocabulary}
\label{subsec:vocab}

FP keeps its core semantics small by reusing the same handful of nouns across planes. In the reference model, every interaction can be described through seven objects: \emph{Entity}, \emph{Session}, \emph{Activity}, \emph{Envelope}, \emph{Event}, \emph{Receipt/Settlement}, and \emph{Provenance}. The vocabulary is intentionally generic. It is rich enough to express tool calls, multi-agent collaboration, organizational workflows, and commerce, yet small enough to remain stable as higher-level patterns evolve. Table~\ref{tab:vocab} summarizes the seven objects. 

\begin{table}[ht]
\centering
\caption{FP core vocabulary in the reference model.}
\label{tab:vocab}
\footnotesize
\begin{tabular}{@{}p{0.22\textwidth} p{0.72\textwidth}@{}}
\toprule
\textbf{Object} & \textbf{Role in FP} \\
\midrule
Entity & An addressable participant, such as an agent, human, tool, resource, service, institution, or organization, with identity, capability metadata, and privacy controls. \\
Session & A scoped multi-party context that binds participants, roles, policy references, and optional budgets, such as spend limits or token ceilings. \\
Activity & A unit of work executed within a session, with explicit state transitions and typed inputs or outputs referenced by context pointers. \\
Envelope & A signed wrapper that carries intent, routing and correlation identifiers, policy references, and minimal metadata for progressive disclosure. \\
Event & An append-only observation emitted during a session or activity, supporting streaming, replay, and backpressure. \\
Receipt / Settlement & Verifiable records of metered usage and value exchange, with references to external payment rails and dispute signals. \\
Provenance & Structured records of decisions and evidence, such as policy outcomes, approvals, and revocations, that make the trace auditable. \\
\bottomrule
\end{tabular}
\end{table}

\subsection{Entity \& Trust Plane}
\label{subsec:entity}

FP begins with a unified entity model. Any participant that can act, be invoked, hold authority, or become part of an interaction is addressable. Each entity exposes four kinds of information: \emph{who} it is (identifiers, keys, and versioning), \emph{what} it can do (capability statements), what \emph{trust signals} others may rely on (attestations, reputation hooks, or other credit signals), and what \emph{privacy} controls govern access and delegation.

A practical design constraint is \emph{overhead}. FP therefore favors progressive disclosure. Capability statements begin as short summaries. A summary may include the entity's purpose, a few risk tags, schema hashes, or hints about pricing and policy. More detail appears only after a counterparty is selected or authorized. Full schemas, examples, and pricing terms can then be fetched by reference. This reduces token usage and avoids the common pattern of copying large tool specifications into a model's working context before they are needed.

Entity identity is also the unit of \emph{accountability} in FP. Organizations can be represented as entities with their own keys and policies. It can hold assets, sponsor sessions, and act as counterparties. Membership becomes a first-class edge with scoped delegation, rather than an application-specific convention. FP still does not require one identity scheme. A deployment may use DIDs, WebPKI, or enterprise identity systems. The protocol only makes the basic structure explicit so the other planes can rely on it.

Trust is treated in the same spirit. FP does not define a global reputation system. Instead, it provides hooks for trust signals, such as attestations, stakes, reputation providers, and policy checks over those signals. This lets deployments begin with local trust and gradually interoperate across domains without reducing trust decisions to ad hoc prompt instructions or private application logic.

\subsection{Transport \& Routing Plane}
\label{subsec:transport}

FP is transport-agnostic by design. The standard defines what message delivery must preserve, but it does not choose the transport. It covers addressing, discovery hooks, channel setup, termination, and flow control. Concrete bindings belong to profiles. This keeps the protocol resilient to changes in network stacks and deployment environments, from local IPC to web-native transports and long-running asynchronous channels.

Routing matters because agentic interactions rarely remain point-to-point. A group session may span several transports at once: local IPC or stdio to a tool, HTTP or WebSocket to a remote agent, and SSE or streaming HTTP to a user interface. FP therefore treats transport as a binding beneath a consistent addressing, correlation, and trace layer. Messages can move across different channels while preserving ordering where required, backpressure, termination semantics, and a coherent record of the interaction. This is a basic requirement for scaling from a handful of cooperating agents to large networks and organizations without losing observability.

\subsection{Interaction \& Organization Plane}
\label{subsec:interaction}

The interaction plane provides the primitives through which entities do things together. \emph{Schemas} define the structure of messages and codecs. \emph{Events and streams} provide ordering, correlation, replay, and backpressure. \emph{Sessions and organizations} capture groups, roles, membership, and delegation as first-class objects. and \emph{Economic primitives} standardize metering, receipts, settlement references, and dispute signals.

A \emph{session} is an explicit container for multi-party collaboration. It binds participants, roles, policy references, and optional budgets such as spend limits or token ceilings. This makes group interaction legible: a bidder in an auction, a reviewer in a regulated workflow, and a tool provider in a pipeline are all represented as roles within a session rather than as application-specific special cases.

Events and streams form the trace layer. FP does not treat collaboration as one long chat transcript. Systems emit typed events instead. Agents can read them. Operators can inspect them. User interfaces can render them. Auditors can review them later.
Backpressure matters once sessions grow. A slow consumer should not have to ingest every event at the same speed as a fast one. It needs control over production, delivery, and replay. Without that control, the trace becomes either incomplete or too expensive to follow. FP keeps collaboration observable without forcing every participant to consume the stream in the same way.


\subsection{Regulation \& Oversight Plane}
\label{subsec:oversight}

FP treats safety as a protocol concern rather than an application afterthought. The oversight plane provides a common place for policy evaluation, enforcement decisions, audit and provenance records, monitoring signals, compliance hooks, and dispute escalation. This aligns with an emerging economic reality: as autonomous systems scale, verification and accountability become scarce resources, and systems that produce low-cost evidence will be easier to deploy, govern, and trust \cite{agi_economics}. In FP, critical decisions can be checked at protocol boundaries, such as before invocation or settlement, and the resulting evidence can be validated by third parties without exposing sensitive payloads.

The oversight plane is intentionally decoupled from any single organization. A deployment may run policy locally, delegate checks to a compliance service, or provide evidence to an external auditor or regulator. FP supports this by allowing policies and provenance records to be referenced, hashed, and verified independently of the payloads they govern. This makes audit portable. The same interaction trace can be inspected under different policies without replaying the interaction.

Oversight also covers failure cases. Disputes, revocations, and safety reports are first-class events. Networks can propagate trust-relevant information through explicit channels instead of relying on informal warnings, private logs, or prompt-level conventions.

\subsection{Configuration \& Profiles}
\label{subsec:profiles}

To avoid an overgrown core, FP makes variability explicit. Profiles bind the core semantics to concrete transports, identity methods, and deployment environments. Registries publish schema and event-type catalogs. Pattern libraries describe reusable multi-party interaction templates, such as auctions, workflows, and bargaining. Bridges adapt existing ecosystems to FP's envelope, trace, policy, and evidence model. This is how FP remains compatible with the protocols in Table~\ref{tab:comparison}: it does not demand replacement, but provides a common control surface across heterogeneous stacks.

This separation also clarifies what belongs in FP and what belongs around it. The core defines the objects and interaction semantics that must work across domains. Profiles choose wire formats and transport bindings. Extensions add event types or interaction patterns. Bridges map external protocols into FP activities. This boundary keeps implementations lightweight and helps the protocol avoid becoming a monolith.

\subsection{Design Principles}
\label{subsec:principles}

The architecture in Figure~\ref{fig:fp-arch} is shaped by a small set of design principles. These principles are not features in the usual sense. They are constraints that keep FP useful as a foundation layer rather than turning it into another application framework.

\textbf{Keep the core small and semantic.} FP standardizes only the objects that must interoperate across domains: entities, sessions and organizations, eventful interaction, and evidence hooks. Anything that can live in a profile, such as transport bindings or identity methods, or in an extension, such as auction schemas or workflow catalogs, is kept out of the core. This keeps the protocol stable as application-level patterns evolve.

\textbf{Prefer progressive disclosure to prompt stuffing.} Many current integrations pay an avoidable cost by copying rich tool descriptions and large capability payloads into the model's working context before they are needed. FP assumes that most interactions should begin with lightweight metadata and become detailed only after selection and authorization. This reduces token overhead, makes caching practical, and keeps the default interaction surface safer.

\textbf{Make multi-party interaction explicit.} Agent systems rarely remain isolated or single-threaded. Even simple deployments quickly involve a personal agent, external tools, service providers, other agents, and human checkpoints. FP therefore makes sessions, roles, membership, and delegation first-class protocol objects, so that group semantics do not have to be simulated in application code. This is also a precondition for larger networks in which coordination structures can form across organizational boundaries.

\textbf{Treat evidence as a first-class output.} FP assumes that execution can easily outpace verification. Policies, audit records, and provenance hooks are therefore part of the protocol's normal flows, not an external logging layer. This does not mean that every payload is recorded. It means that important decisions can be explained and validated later, including by parties that were not present at execution time \cite{agi_economics}.

These principles make FP strict in some places and flexible in others. FP is strict about the parts that must stay coherent across systems: entities, organizations, evidence, and policy. It is flexible about the parts that should evolve outside the standard, such as transports, identity mechanisms, vertical patterns, and bridges. This separation lets FP function as infrastructure rather than as a one-off integration recipe.
In practice, FP can be adopted around existing systems. Teams can begin by wrapping a small number of tools or agents to gain consistent identity, traces, and policy enforcement, then progressively add organization and economic primitives as their deployments mature.

\section{Application Scenarios}
\label{sec:scenarios}

Protocols matter when they make difficult coordination feel ordinary. The previous section described FP's architecture. This section turns to the kinds of systems that architecture is meant to support, and then follows one concrete scenario in which all planes of the protocol become active.

\subsection{Where FP applies}
\label{subsec:scenario-overview}

FP is built around a simple observation. As agents become more capable, they begin to face coordination problems that human organizations have long managed through roles, contracts, budgets, audits, and markets. These problems do not appear in only one domain. They recur whenever autonomous entities must discover one another, delegate work, exchange resources, enforce policy, and leave evidence that can be inspected later.

One recurring case is cross-protocol interoperation. A single workflow may combine tool invocation through MCP, agent-to-agent delegation through A2A, and controllable user-interface delegation through A2UI \cite{mcp_spec,a2a_spec,a2ui_spec}. Without a foundation layer, each path brings its own notion of identity, session state, authority, and logging. FP provides a shared substrate of entities, sessions, envelopes, and traces, so that heterogeneous interactions can compose over one control surface.

A second case is organization and governance. As agents specialize, the unit of coordination shifts from a single assistant to a team, service network, or institution. FP treats organizations, roles, membership, and delegation as protocol objects. Policy enforcement points can gate sensitive transitions, while a shared evidence spine supports internal review and external compliance \cite{agi_economics,virtual_agent_economies}.

Economic activity is another natural setting. Service procurement, resource allocation, and marketplace interaction require metering, attestation, settlement, and dispute handling. FP represents these as ledger-agnostic primitives, so that economic relationships can be audited without requiring a particular payment rail \cite{ucp_spec}.

The same logic also applies to social coordination and collective governance. Open agent networks can amplify useful coordination, but they can also amplify manipulation, spam, and instruction injection. FP makes communities representable as organizations with explicit moderation roles, policy hooks, and provenance signals. This turns social governance into a protocol-level capability rather than an application-specific afterthought \cite{moltbook_site,wired_moltbook,microsoft_openclaw_security}.

Finally, FP is especially relevant when autonomy must coexist with supervision. In domains such as healthcare, finance, research, and government, systems need ways to limit unnecessary data exposure, require human approval for sensitive transitions, and preserve provenance records that auditors can later inspect. These categories are not independent. Real deployments usually combine several of them. The following scenario makes this combination concrete.

\subsection{A concrete example: an AI company with human oversight}
\label{subsec:concrete-example}

Consider a small AI company created by a human founder to deliver a software product. The company is not a traditional firm with a large staff. It is an organization of specialized agents, external tools, service providers, and human checkpoints. It must discover and hire outside capabilities, coordinate heterogeneous participants, manage budgets, and remain auditable as work proceeds. The example is useful because it brings every FP plane into view without requiring a contrived setting.

\begin{figure}[!ht]
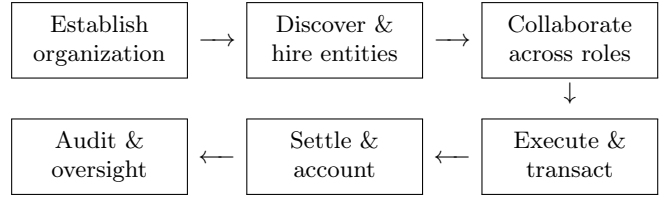

    \centering
    \small
    \setlength{\tabcolsep}{0pt}
    \begin{tabular}{c@{\hspace{4pt}}c@{\hspace{4pt}}c@{\hspace{4pt}}c@{\hspace{4pt}}c@{\hspace{4pt}}c}
    \fbox{\parbox{2.1cm}{\centering\strut Establish\\organization\strut}}
    & $\xrightarrow{\hspace{8pt}}$
    & \fbox{\parbox{2.1cm}{\centering\strut Discover \&\\hire entities\strut}}
    & $\xrightarrow{\hspace{8pt}}$
    & \fbox{\parbox{2.1cm}{\centering\strut Collaborate\\across roles\strut}}
    & \\[8pt]
    & & & & $\downarrow$ & \\[4pt]
    \fbox{\parbox{2.1cm}{\centering\strut Audit \&\\oversight\strut}}
    & $\xleftarrow{\hspace{8pt}}$
    & \fbox{\parbox{2.1cm}{\centering\strut Settle \&\\account\strut}}
    & $\xleftarrow{\hspace{8pt}}$
    & \fbox{\parbox{2.1cm}{\centering\strut Execute \&\\transact\strut}}
    & \\
    \end{tabular}
    \caption{Lifecycle of the AI company scenario. Each phase maps to one or more FP planes. The flow is representative rather than strictly sequential, since phases can overlap and repeat as the organization evolves.}
    \label{fig:scenario-flow}
\end{figure}

Figure~\ref{fig:scenario-flow} sketches the lifecycle. The phases are presented in order for clarity, although in practice they can overlap and repeat as the organization evolves.

\subsubsection{Phase 1: Establishing the organization}

The founder registers as an Entity on a local Host and creates an organization, which is also represented as an Entity, with its own governance policy. She then registers three internal agent entities: a \emph{planner} for task decomposition, a \emph{developer} for code generation and testing, and a \emph{reviewer} for quality assurance. Each receives an FP address, a cryptographic identity, and a role inside the organization.

This phase exercises the Entity \& Trust Plane and the Interaction \& Organization Plane. The unified entity model gives humans, agents, tools, services, and organizations the same addressing and identity surface. At the same time, roles, membership, and approval rules become protocol objects rather than application-level conventions. A rule such as ``deployments require human approval'' can therefore be attached to the organization and enforced at protocol boundaries.

\subsubsection{Phase 2: Discovering and hiring external entities}

The planner needs a GPU provider for training and an MCP-compatible code-search tool for development. It queries the network's discovery mechanism, which aggregates publicly discoverable EntityCards from the local Host, its children, and its parent. Discovery returns lightweight metadata, such as name, kind, capability summary, and pricing hints, without exposing full schemas.

After comparing candidates, the organization engages the GPU provider and the code-search tool by forming trust relationships and establishing sessions with scoped roles and budget limits. The external entities are now addressable within FP's routing graph and remain subject to the organization's access-control policies.

Here the Transport \& Routing Plane provides discovery and reachability, while progressive disclosure keeps the evaluation phase lightweight. The Entity \& Trust Plane supplies the identity basis through EntityCard exchange, and the Interaction \& Organization Plane turns a service relationship into an inspectable session with roles and budget ceilings.

\subsubsection{Phase 3: Heterogeneous group collaboration}

The founder assigns a task: ``ship the authentication module.'' The planner decomposes the task and distributes the subtasks. The developer writes code and invokes the external code-search tool through FP's MCP bridge. The reviewer inspects the output. Before deployment, the founder receives a human-approval request.

These interactions involve humans, agents, and tools, but they do not require separate coordination mechanisms. They pass through the same envelope, routing, and checkpoint infrastructure. Messages are signed, routed through the host tree, and validated by the checkpoint pipeline before reaching their handlers. The session maintains one trace across all participants, preserving correlation and causal links.

This phase mainly exercises the Interaction \& Organization Plane, the Transport \& Routing Plane, and Configuration \& Profiles. Sessions, activities, and event streams coordinate the work. Routing connects local and remote entities, with offline queues handling transient disconnections. Bridges allow external providers and MCP-compatible tools to appear on the FP surface without changing their original protocols.

\subsubsection{Phase 4: Economic activity}

The GPU provider reports metered usage, such as compute-hours and token consumption, as structured records attached to the session's activities. When the work is complete, it issues a signed receipt that attests to what was delivered and references an external payment rail for settlement. The organization's budget ceiling, set when the session was created, is enforced throughout the interaction. A payment checkpoint in the access-control pipeline can reject messages that would exceed the authorized spend.

This phase brings together economic primitives and oversight. Metering, receipts, settlement references, and dispute signals represent the transaction lifecycle within the protocol. Budget policy is enforced at a checkpoint, and receipts can be verified by third parties without exposing sensitive payloads.

\subsubsection{Phase 5: Oversight and audit}

Throughout the lifecycle, policy decisions leave structured traces. The reviewer's approval, the founder's deployment authorization, the payment checkpoint's budget decision, and any rejected messages are recorded as events. Provenance records bind these decisions to the policies and evidence that produced them, creating an audit trail that can be inspected later by parties who were not present during execution.

When the founder reviews the completed project, she can ask which entity performed each action, under which policy, with what economic outcome, and whether any access-control decisions were overridden. If a dispute arises with the GPU provider, the same evidence spine supports resolution without reconstructing state from scattered logs.

This final phase shows why Entity \& Trust and Regulation \& Oversight must work together. Signatures on envelopes help make records tamper-evident, while policy enforcement points, provenance records, and dispute signals make the lifecycle auditable.

\begin{table}[!ht]
\centering
\caption{How the AI company scenario exercises the FP planes.}
\label{tab:scenario-planes}
\footnotesize
\begin{tabular}{@{}p{0.24\textwidth} p{0.32\textwidth} p{0.36\textwidth}@{}}
\toprule
\textbf{Phase} & \textbf{Main FP planes} & \textbf{Protocol role} \\
\midrule
Establish organization & Entity \& Trust; Interaction \& Organization & Register entities, assign roles, and attach governance policy. \\
Discover and hire entities & Transport \& Routing; Entity \& Trust; Interaction \& Organization & Discover capabilities, verify identity, and create scoped sessions. \\
Collaborate across roles & Interaction \& Organization; Transport \& Routing; Configuration \& Profiles & Coordinate humans, agents, and tools through shared envelopes and bridges. \\
Execute and transact & Interaction \& Organization; Regulation \& Oversight & Meter usage, issue receipts, and enforce budget policy. \\
Audit and oversight & Regulation \& Oversight; Entity \& Trust & Preserve provenance, validate decisions, and support dispute resolution. \\
\bottomrule
\end{tabular}
\end{table}

\subsection{What the example shows}
\label{subsec:scenario-takeaway}

The scenario exercises all five planes in FP's architecture. The Entity \& Trust Plane handles registration and identity. The Transport \& Routing Plane handles discovery and message routing. The Interaction \& Organization Plane handles organizational structure, sessions, and economic primitives. The Regulation \& Oversight Plane handles policy enforcement and audit. Configuration \& Profiles handle the protocol bridges.

The important point is not that the scenario requires many special mechanisms. It is the opposite. The same entity model that represents the founder also represents an external GPU provider. The same checkpoint pipeline that enforces access control can enforce budget limits. The same envelope that carries a task message can also carry a payment receipt. This uniformity is the practical consequence of FP's design: a small protocol core that composes, rather than a collection of point solutions that must later be stitched together.

\section{Conclusion}
\label{sec:conclusion}

Autonomous agents are becoming social and economic actors. As they scale, the main constraints shift from isolated capability to coordination, governance, and evidence. FP proposes a compact protocol core for this emerging agentic society: a unified entity model, native organization primitives, eventful interaction, ledger-agnostic economic attestation, and protocol-level oversight. By separating a small core from profiles, extensions, and bridges, FP is designed to complement existing protocols while providing a stable substrate for large-scale, trustworthy cooperation.

An open and governable agentic society cannot be built from isolated point solutions alone. It needs a communication substrate in which entities, organizations, value attestations, and evidence remain interoperable across tools, agents, humans, services, and institutions. FP is a step toward that substrate. Its purpose is to make autonomous agency composable without making accountability optional. The next step is to turn this architecture into a precise specification and a set of reference bindings, so that the ecosystem can evaluate, implement, and refine FP in practice.

\section*{Acknowledge}
\label{sec:ack}
Bang Liu gratefully acknowledges the funding support from the Amazon Research Award (ARA) program and the Canada CIFAR AI Chair program.

\bibliographystyle{plain}
\bibliography{main}

\newpage
\appendix

\section{Reference Implementation}
\label{app:ref}

The Foundation Protocol is accompanied by an open-source reference stack:
the \texttt{foundation-protocol} runtime\footnote{
\url{https://github.com/FoundationAgents/foundation-protocol}}
and the \texttt{ai-link-net} application-network implementation\footnote{
\url{https://github.com/FoundationAgents/ai-link-net}}.
Both are non-normative. The main text defines the protocol. This appendix describes their architecture and main technical choices, so readers can evaluate the design's feasibility and trade-offs.
Implementation details that only make sense with the code, such as module names, file paths, and API signatures, are deliberately omitted. The goal is a software-architecture blueprint that can stand on its own.

\subsection{Overview}
\label{app:overview}

The two released repositories form a working FP stack. \texttt{foundation-protocol} contains the protocol core and the Python runtime. \texttt{ai-link-net} builds on it with an application-network server, a command-line interface, and a web interface.
The stack supports the basic FP workflow. Humans, agents, tools, resources, and services can register on a Host. They can discover peers across the network, form trust relationships, and exchange signed and encrypted messages. They interact through the same protocol surface instead of separate application paths.

The core, server, and command-line interface are written in Python. The web interface uses TypeScript, React, and Vite. The implementation targets developer-facing use cases first. A developer can stand up a local or cloud-hosted FP node, register entities, and run cross-host interactions with real AI providers such as Claude Code and Codex CLI, as well as MCP-compatible tool servers.

\subsection{Core concepts}
\label{app:concepts}

Three abstractions form the structural backbone of the implementation.

\paragraph{Entity.}
An Entity is the unified participant model introduced in Section~\ref{subsec:entity}. FP represents every addressable participant as an Entity. This includes human users, AI agents, tools, resources, services, and organizations. Each Entity uses the same surface for identity, addressing, cryptography, and access control.
FP keeps this abstraction deliberately flat. Hosts, humans, agents, tools, resources, services, arbiters, and organizations all share the same Entity model. Behavioral differences live in pluggable \emph{handlers}, not in subtype hierarchies. A new entity kind therefore needs a new handler. It does not need a new protocol path.

Each Entity is assigned a globally unique address of the form
\texttt{HostUid:EntityUid}, where both components are short hex strings.
An Entity publishes an \emph{EntityCard}, a lightweight discovery
document containing its name, address, kind, and public keys. Counterparts can use it to verify identity and establish encrypted channels. They do not need to exchange full capability metadata upfront. This follows FP's progressive-disclosure principle.

\paragraph{Host.}
A Host is the topology and routing node in FP. It maintains a registry of local entities. It manages parent--child relationships with other Hosts. It also routes mail to the right destination: a local entity, a child Host, or a parent Host.
The Host is still a protocol object, not an implementation recipe. It defines addressing, discovery, and routing semantics. It does not prescribe how a deployment maintains connections or persists state.

Discovery relies on a well-known endpoint that each Host exposes,
returning its identity, URL, and the set of publicly discoverable
entities. When participants aggregate these documents across a host tree, they get a view of the reachable network.

\paragraph{Server.}
A Server is the application-layer runtime around a Host. It connects the Host's abstract routing model to real network connections. In the reference implementation, the Server binds routing to WebSocket connections.
It tracks entity presence, queues offline mail with bounded buffers, and uses heartbeats to detect liveness. It also exposes an HTTP REST API for external clients.
This keeps the boundary clean. The Host defines protocol semantics. The Server handles runtime concerns. As a result, the protocol remains easier to test and easier to port across web frameworks.
\medskip
\noindent This decomposition between Entity, Host, and
Server reflects that protocol semantics are
isolated from runtime concerns, and both are isolated from interface presentation. 

Because of this, the protocol core can be embedded in contexts that have nothing to do with HTTP, e.g., batch scripts, test harnesses, alternative transport layers.

\subsection{Architecture}
\label{app:architecture}

\subsubsection{Layering and dependency direction}

The codebase is organized into three layers with a strict unidirectional
dependency rule:

\begin{itemize}
    \item \textbf{Protocol core.}  Entity model, message semantics, mail
      envelope, cryptography, host topology, checkpoint pipeline, handler
      abstractions, and protocol adapters (e.g., CLI adapter for AI
      providers, MCP bridge for tool servers).  This layer has no
      dependency on any web framework, database, or persistence backend.
      It is a pure domain model.

    \item \textbf{Application layer.}  HTTP/WebSocket server, runtime
      host management, client libraries for host-to-host communication,
      API schemas, and configuration management.

    \item \textbf{Interface layers.}  CLI commands and a React web~UI.
      Both consume the application layer's API; neither contains protocol
      logic.
\end{itemize}

\noindent The dependency direction is strictly
\emph{application~$\rightarrow$~core}, i.e., the protocol core never imports
from the application layer.  This constraint ensures that protocol
semantics remain stable and testable independently of deployment choices.

\subsubsection{Concurrency model}

All I/O paths in the server are fully asynchronous.  Entity message
processing, host-to-host forwarding, WebSocket management, and offline
queue flushing run as concurrent coroutines on a single event loop.
This design supports high-fanout scenarios, where a single Host routes
mail among many entities and child Hosts simultaneously, without
blocking on individual message deliveries.

\subsubsection{Topology: tree-based with a decentralization path}

The current implementation uses a \textbf{tree topology}. Each Host has at most one parent and may have many children. A cloud-hosted root node can act as a rendezvous point. Local Hosts then connect as children.
This gives FP a simple operational default. It covers a common deployment pattern where a personal or team Host joins a shared network through a well-known entry point.

\begin{quote}
\small\ttfamily
\begin{tabular}{@{}l@{}}
\hspace{12em}CloudHost (root) \\
\hspace{8em}/\hspace{8em}\textbackslash \\
\hspace{2em}LocalHost~A\hspace{8em}LocalHost~B \\
\hspace{1em}Alice~~DevBot\hspace{8em}Bob~~EchoTool \\
\end{tabular}
\end{quote}

The tree model is not a hard architectural constraint.  The routing
algorithm dispatches based on Host~UID lookup: local, child, or parent.
A peer-to-peer or mesh topology would add lateral routing entries. That change stays in the routing layer. It does not require changes to the entity model, mail format, checkpoint pipeline, or other protocol components.
FP is designed so topology can evolve without forcing protocol-level changes.

\subsubsection{Transport independence}

The protocol core defines addressing and routing semantics but does not
bind to a specific transport.  The current default profile uses WebSocket
for persistent host-to-host connections (with heartbeat, reconnection,
and offline queuing) and HTTP~REST for client-to-host operations.  These
bindings live entirely in the application layer.  Alternative transports
(QUIC, gRPC, local IPC) can be introduced as additional profiles without
modifying the protocol core, consistent with FP's configuration and
profile plane (Section~\ref{subsec:profiles}).

\subsection{Core functionality: technical overview}
\label{app:functionality}

\subsubsection{Unified entity model and access control}

All core entity kinds share the same addressing scheme, the same
cryptographic identity (signing keys and encryption keys),
and the same access-control surface.  This uniformity means that an
agent invoking a tool, a human messaging another human, and a service
responding to a resource request all traverse identical envelope,
routing, and policy mechanisms.

Access control uses a \emph{checkpoint pipeline}. Every inbound message passes through an ordered sequence of policy enforcement points before it reaches a handler. Checkpoints are pluggable and composable. The reference implementation includes checkpoints for:

\begin{itemize}
    \item Friend-list-based access control (social-graph gating).
    \item Session validation (messages must reference an active session).
    \item Rate limiting (sliding-window, per-sender).
    \item Content-length enforcement.
    \item Payment verification (reject messages lacking a valid payment proof).
    \item Contract approval (intercept contract status changes that require owner decision).
    \item Payment approval (evaluate payment requests against configurable policies).
\end{itemize}

\noindent Additional checkpoints can be appended without modifying
existing ones, and their evaluation order determines priority.  This
pipeline is the implementation-level realization of the policy
enforcement points described in the Regulation \& Oversight Plane
(Section~\ref{subsec:oversight}).

\paragraph{Human-in-the-loop approval.}
The checkpoint pipeline supports a general-purpose \emph{owner
escalation} mechanism. Any checkpoint can defer a decision to the entity's designated owner, usually a human. The checkpoint sends an approval request. The original message waits in a pending queue. Processing continues only after the owner explicitly approves or rejects the request.
The trade subsystem uses this mechanism for contract creation, delivery acceptance, and high-value payments. Custom checkpoints can use it as well. Human oversight is therefore not a wrapper around the protocol. It is part of the checkpoint pipeline itself, with the same ordering, tracing, and audit guarantees as automated checks.

Behavioral specialization across entity kinds is handled by a
\emph{handler} abstraction.  Each entity is assigned a handler that
determines how incoming messages are processed after passing all
checkpoints.  The reference implementation provides handlers for human
entities (store-and-forward for UI consumption), agent entities (delegate
to an external AI provider via a CLI adapter), tool/resource/service
entities (bridge to an MCP server via JSON-RPC), and a generic callback
handler for programmatic use. The handler interface is the primary
extension point for adding new entity behaviors.

\subsubsection{Protocol bridges}

Two bridges demonstrate FP's ability to wrap existing ecosystems:

\begin{itemize}
    \item \textbf{CLI adapter bridge.}  Agent entities delegate work to
      external AI providers (e.g., Claude Code, Codex~CLI) through a
      declarative configuration that maps FP's provider-agnostic
      execution semantics (trust level, output format, budget, allowed
      tools) to provider-specific CLI flags.  Adding a new provider
      requires only a configuration entry, not code changes.  The adapter
      manages subprocess lifecycle, output parsing, and session
      resumption.

    \item \textbf{MCP bridge.}  Tool, resource, and service entities can
      front any MCP-compatible server.  The bridge translates FP
      \texttt{INVOKE} messages into JSON-RPC \texttt{tools/call} requests
      and returns results as FP messages.  Both STDIO and HTTP transports
      are supported.  This means an existing MCP server becomes an
      FP-addressable entity, with identity, friend-based access control,
      and mail-based invocation, without any modification to the server
      itself.
\end{itemize}

\noindent These bridges illustrate the design objective stated in
Section~\ref{subsec:objectives}. FP complements existing protocols by
providing a common control surface, rather than replacing their
domain-specific semantics.

\subsubsection{Contract lifecycle and economic functionality}

The reference implementation realizes FP's economy primitives through a complete \emph{contract-and-settlement} subsystem. The subsystem has three protocol roles. The buyer and seller act as the two \emph{parties}. An \emph{arbiter} manages state transitions, signs attestations, and can optionally hold funds.

\paragraph{Contract state machine.}
A contract follows a well-defined lifecycle:

\begin{quote}
\small\ttfamily
DRAFT $\rightarrow$ PENDING $\rightarrow$ ACTIVE $\rightarrow$ COMPLETING $\rightarrow$ SETTLING $\rightarrow$ SETTLED \\
\quad\, $\downarrow$ \hspace{4.2em} $\downarrow$ \hspace{3.8em} $\downarrow$ \hspace{5.6em} $\downarrow$ \\
\quad CANCELLED \hspace{1em} CANCELLED \hspace{0.8em} DISPUTED \hspace{2em} DISPUTED
\end{quote}

\noindent Each transition is triggered by a typed message, such as \texttt{CONTRACT\_APPROVE}, \texttt{CONTRACT\_COMPLETE}, or \texttt{CONTRACT\_ACCEPT}. The arbiter validates the transition before it takes effect. During the draft phase, parties can still amend the terms. Each amendment resets approvals and increments a draft version counter. This prevents stale-state races. A configurable rework limit controls how many times a delivery may be revised before the contract escalates into a dispute.

\paragraph{Payment and settlement.}
The payment subsystem supports two funding modes. In \emph{escrow} mode, the arbiter freezes funds from the buyer when the contract becomes active. It transfers the funds to the seller after successful settlement. In this mode, the protocol itself guarantees atomicity.
In \emph{direct} mode, the parties settle outside FP through a bank transfer, cryptocurrency payment, or external payment gateway. The arbiter then records the settlement as a verifiable reference. Both modes share the same receipt and dispute surface, so downstream audit does not depend on which payment rail was used.

\paragraph{Metering and receipts.}
Each delivery in a contract carries structured cost records. These may include token consumption, compute-hours, or USD amounts. FP attaches the records to the contract's activity history. After acceptance, the arbiter issues a signed receipt. The receipt attests to what was delivered and what it cost.
These receipts are third-party verifiable. An auditor can confirm a receipt's integrity with the arbiter's public key without accessing the underlying payloads.

\paragraph{Reputation.}
The implementation derives \emph{explainable vendor reputation} from
closed-loop contract facts rather than subjective ratings.  When a
contract reaches a terminal state (settled, cancelled, or disputed), the
system extracts a reputation event that records the outcome, delivery
count, rework count, cost data, and evidence completeness.  These events
are aggregated into a five-dimensional profile.

\begin{itemize}[nosep]
    \item \emph{Quality}: inferred from explicit ratings and outcome
      type.
    \item \emph{Reliability}: proportion of contracts completed
      successfully.
    \item \emph{Collaboration}: penalized by rework frequency.
    \item \emph{Efficiency}: delivery iterations and cost transparency.
    \item \emph{Integrity}: presence of signed snapshots, hash chains,
      and complete evidence.
\end{itemize}

\noindent Each dimension is weighted by a confidence factor (reflecting
sample size and data quality) and a recency factor (recent contracts
weighted higher).  The result is a reputation score that remains tied to concrete contract evidence rather than to an opaque aggregate number. A reviewer can always trace the score back to the contracts and events that produced it.

\subsubsection{Security and auditability}

The security model operates at three levels.

\paragraph{Envelope security.}
Every mail message is signed before it is sent. The recipient checks the signature against the sender's public key from the sender's EntityCard. If the payload needs confidentiality, the sender encrypts it with authenticated encryption.
This gives two guarantees. \emph{Authenticity} means the sender is the expected entity and the message has not changed in transit. \emph{Confidentiality} means only the intended recipient can read the payload.
FP defines these guarantees, but it does not hard-code the algorithms. The application layer chooses them. The reference implementation uses Ed25519 for signatures and ephemeral X25519 key agreement with AES-256-GCM for encryption. A different cryptographic profile can replace these choices without changing the envelope format or the checkpoint pipeline.

\paragraph{Contract-level audit trail.}
The trade subsystem adds a second layer of cryptographic evidence.
Every contract state transition produces an immutable \emph{snapshot}. The arbiter signs the snapshot. Each snapshot also stores the hash of the previous one, so the contract history forms a hash chain. Retroactive edits become detectable.
The system captures participant EntityCards when the contract is created. The audit record therefore remains readable even if an entity later changes its metadata. Delivery artifacts, cost reports, and settlement references all attach to the same chain. An auditor can inspect one evidence spine for the whole contract.

\paragraph{Checkpoint-level traceability.}
Each entity maintains an append-only mailbox that records all inbound and outbound mail in chronological order. 
The checkpoint pipeline also records policy decisions. If a checkpoint rejects a message, the system records the reason and returns a structured error. If a checkpoint asks a human owner to decide, the system records the request, the owner's response, and the action that follows. In other words, access-control decisions leave traces. They do not vanish as silent drops.

\end{document}